\documentclass{article} 
\usepackage{iclr2024_conference,times}

\usepackage{amsmath,amsfonts,bm}

\def\eqref#1{equation~\ref{#1}}

\DeclareMathAlphabet{\mathsfit}{\encodingdefault}{\sfdefault}{m}{sl}
\SetMathAlphabet{\mathsfit}{bold}{\encodingdefault}{\sfdefault}{bx}{n}

\usepackage{xcolor}
\definecolor{link-color}{RGB}{25,33,205}
\definecolor{dark-blue}{RGB}{4,13,112}
\usepackage[colorlinks,linkcolor=dark-blue,urlcolor=link-color,citecolor=dark-blue]{hyperref}
\usepackage{url}
\usepackage{cleveref}
\usepackage{outlines}
\usepackage{multirow}

\usepackage{booktabs}       
\usepackage{amsfonts}       
\usepackage{nicefrac}       
\usepackage{microtype}      
\usepackage{graphicx}
\usepackage{scalerel}
\usepackage{pifont}
\usepackage{dsfont}
\usepackage{amssymb}
\usepackage{wrapfig}

\usepackage{caption}
\usepackage{subcaption}
\usepackage{wrapfig}
\usepackage[ruled]{algorithm2e}
\usepackage{adjustbox}

\newcommand{\mask}{\mathbf{m_n}}

\newcommand{\posss}{\mathbf{x}}

\newcommand{\dirv}{\mathbf{d}}
\newcommand{\colorv}{\mathbf{c}}
\newcommand{\density}{\sigma}
\newcommand{\F}{F_{\Phi}}
\newcommand{\Fdensity}{\F^{\density}}
\newcommand{\Fedit}{\hat{\F}}
\newcommand{\boundary}{\mathbf{V_o}}
\newcommand{\transform}{R_o}
\newcommand{\transformK}[1]{\transform^{#1}}
\newcommand{\graph}{G}
\newcommand{\graphT}[1]{\graph^{#1}}
\newcommand{\kinematic}{a}
\newcommand{\boundaryShift}{\Delta \mathbf{x}_\boundary}
\newcommand{\physprops}{\mathbf{p}}

\newcommand{\movicrolloutone}{031}
\newcommand{\movicrollouttwo}{034}
\newcommand{\movicrolloutthree}{046}

\newcommand{\FIGNOMESHMOVIC}{FIG-no-mesh_xid74908777_wid1}

\newcommand{\MOVICROW}[1]{
    \begin{subfigure}[b]{0.1\textwidth}
    \centering
     Ground\\truth\vspace{1cm}
    \end{subfigure}
    \begin{subfigure}[b]{0.89\textwidth}
    \adjincludegraphics[width=\textwidth,trim={0 0 0 {.5\height}},clip]{figures/rollouts/movic_traj#1_\FIGNOMESHMOVIC.png}
    \end{subfigure}\\
    \begin{subfigure}[b]{0.1\textwidth}
     \centering
     FIGNet*\vspace{1cm}
    \end{subfigure}
    \begin{subfigure}[b]{0.89\textwidth}
    \adjincludegraphics[width=\textwidth,trim={0 {.5\height} 0 0},clip]{figures/rollouts/movic_traj#1_\FIGNOMESHMOVIC.png}
    \end{subfigure}\\
}

\title{Scaling Face Interaction Graph Networks to Real World Scenes}

\author{Tatiana Lopez-Guevara, ~Yulia Rubanova, ~William F. Whitney, ~Tobias Pfaff, \\ \textbf{Kimberly Stachenfeld, Kelsey R. Allen} \\
Google DeepMind \\
}

\iclrfinalcopy 
\begin{document}

\maketitle

\begin{abstract}
Accurately simulating real world object dynamics is essential for various applications such as robotics, engineering, graphics, and design. 
To better capture complex real dynamics such as contact and friction, learned simulators based on graph networks have recently shown great promise \citep{allen2023fig,allen2022graph}. However, applying these learned simulators to real scenes comes with two major challenges: first, scaling learned simulators to handle the complexity of real world scenes which can involve hundreds of objects each with complicated 3D shapes, and second, handling inputs from perception rather than 3D state information.
Here we introduce a method which substantially reduces the memory required to run graph-based learned simulators. Based on this memory-efficient simulation model, we then present a perceptual interface in the form of editable NeRFs which can convert real-world scenes into a structured representation that can be processed by graph network simulator.
We show that our method uses substantially less memory than previous graph-based simulators while retaining their accuracy, and that the simulators learned in synthetic environments can be applied to real world scenes captured from multiple camera angles. This paves the way for expanding the application of learned simulators to settings where only perceptual information is available at inference time.
\end{abstract}

\section{Introduction}

Simulating rigid body dynamics is an important but challenging task with broad applications ranging from robotics to graphics to engineering. Widely used analytic rigid body simulators in robotics such as Bullet \citep{Coumans2015}, MuJoCo \citep{todorov2012mujoco}, and Drake \citep{drake} can produce plausible predicted trajectories in simulation, but system identification is not always sufficient to bridge the gap between real world scenes and these simulators~\citep{wieber2016modeling, Stewart1996a, fazeli2017empirical, lan2022affine, parmar2021fundamental, guevara2017adaptable}. This is due, in part, to the challenges of estimating fine-grained surface structures of objects which often have large impacts on their associated dynamics \citep{Bauza2017probabilistic}. This fundamental issue contributes to the well-documented sim-to-real gap between outcomes from analytical solvers and real-world experiments.

Learned simulators have shown the potential to fill the sim-to-real gap \citep{allen2023fig,allen2022graph} by representing rigid body dynamics with graph neural networks. These fully learned simulators can be applied directly to real world object trajectories, and do not assume any analytical form for rigid body contacts. As a result, they can learn to be more accurate than system identification with an analytic simulator even with reasonably few real world trajectories.

However, real world scenes present major challenges for learned simulators. First, learned simulators generally assume access to full state information (the positions, rotations, and exact shapes of all objects) in order to simulate a trajectory. This information must be inferred from a collection of sensor measurements. Second, learned simulators can be memory intensive, especially for the kinds of intricate, irregular objects that often comprise real-world scenes. The currently best-performing graph-based methods operate on explicit surface representations, i.e. point clouds or triangulated meshes~\citep{pfaff2021learning}. The induced graphs of these methods tend to consume vast amounts of GPU memory for complex object geometries, or when there are many objects in the scene. Consequently, results are generally shown for scenes containing fewer than 10 objects with reasonably simple object geometries.


Here we propose a simple, yet surprisingly effective modification (FIGNet*) to the learned, mesh-based FIGNet rigid body simulator~\citep{allen2023fig} that can address these challenges with representing and simulating real world scenes:
\begin{itemize}
    \item FIGNet* consumes much less memory, while maintaining translation and rotation rollout accuracy. This allows us to train FIGNet* on datasets with more objects with complex geometries such as Kubric MOVi-C, which FIGNet cannot train on due to memory cost.
    \item We connect a NeRF perceptual front-end \citep{barron2022mipnerf360} to FIGNet*, and show that we can simulate plausible trajectories for complex, never-before-seen objects in real world scenes.
    \item We show that despite training FIGNet* on simulated rigid body dynamics with ground-truth meshes, the model is robust to noisy mesh estimates obtained from real-world NeRF data.

\end{itemize}

\section{Related work}

\paragraph{Learned simulators} attempt to replicate analytical simulators by employing a learned function approximator. Typically, they are trained using ground truth state information, and consequently cannot be directly applied to visual input data. The representation of state varies depending on the method, but can range from point clouds \citep{Li2018,sanchez2020learning,mrowca2018flexible,linkerhagner2023grounding}, to meshes \citep{pfaff2021learning,allen2023fig}, to signed distance functions (SDFs) \citep{le2023differentiable}. Subsequently, learned function approximators such as multi-layer perceptrons (MLPs) \citep{li2021nerfdy}, graph neural networks (GNNs) \citep{battaglia2018relational,sanchez2018}, or continuous convolutional kernels \citep{ummenhofer2019lagrangian} can be employed to model the temporal evolution of the state. Our approach follows the mesh-based state representation options, but aims to provide a more efficient graph neural network dynamics model.

\paragraph{Bridging simulators to perception.} 
Multiple approaches aim to bridge these learned simulators to perceptual data.  Some approaches are ``end-to-end'' -- they train a perceptual input system jointly with a dynamics model, often assuming access to ground truth state information like object masks \citep{janner2019reasoning,driess2022learning,shi2022robocraft,xue20233dintphys,whitney2023learning}. Others first learn a perceptual encoder and decoder, and then fix these to train a dynamics model in latent space \citep{li2021nerfdy}. 

Most related to our approach are methods that use neural radiance fields to reconstruct 3D scenes from 2D multi-view scenes to enable simulation. Some of these assume hand-crafted but differentiable dynamics models \citep{qiao2023dmrf,qiao2022neuphysics,chu2022physics}, while others learn the dynamics model separately from state information \cite{guan2022neurofluid}. We similarly aim to simply apply our pre-trained learned simulators to real scenes by using a NeRF perceptual front-end. We show that this approach can work \emph{without} fine-tuning even when simulators are trained only from synthetic data.





\section{Method}

\subsection{FIGNet*}
FIGNet* closely follows the method of Face Interaction Graph Networks (FIGNet) \citep{allen2023fig} which is a graph neural network approach designed for modeling rigid body dynamics. In FIGNet, each object is represented as a triangulated mesh $M$ made of triangular mesh faces $\{\mathcal{F}_M \}$ with mesh vertices
$\{\mathcal{V}_M \}$. A scene graph $\mathcal{G}$ then consists of $O$ objects, each with their own triangulated meshes $M_o$. 
At any given time $t$, $M_o^t$ can be represented using the object's transformation matrix, $M_o^t = \transformK{t} \times M_o$.
A simulation trajectory is represented as a sequence of scene graphs $\mathcal{G} = (G^{t_0}, G^{t_1}, G^{t_2}, \dots)$ constructed from these meshes. FIGNet is then a simulator $S$ parameterized by neural network weights $\Theta$, trained to predict the next state of the physical system $\tilde{G}^{t+1}$ based on the previous two scene graphs $\{G^{t}, G^{t-1}\}$, ie $G^{t+1} = S_\Theta(G^t, G^{t-1})$. We train with a mean-squared-error loss on the predicted positions of the vertices for each object $\{\mathcal{V}_M \}$. During inference, $S_\Theta$ can be recursively applied to yield a rollout of any length $T$.

\begin{wrapfigure}{r}{0.35\textwidth}
\centering
\vspace{-2em}
\includegraphics[width=0.33\textwidth]{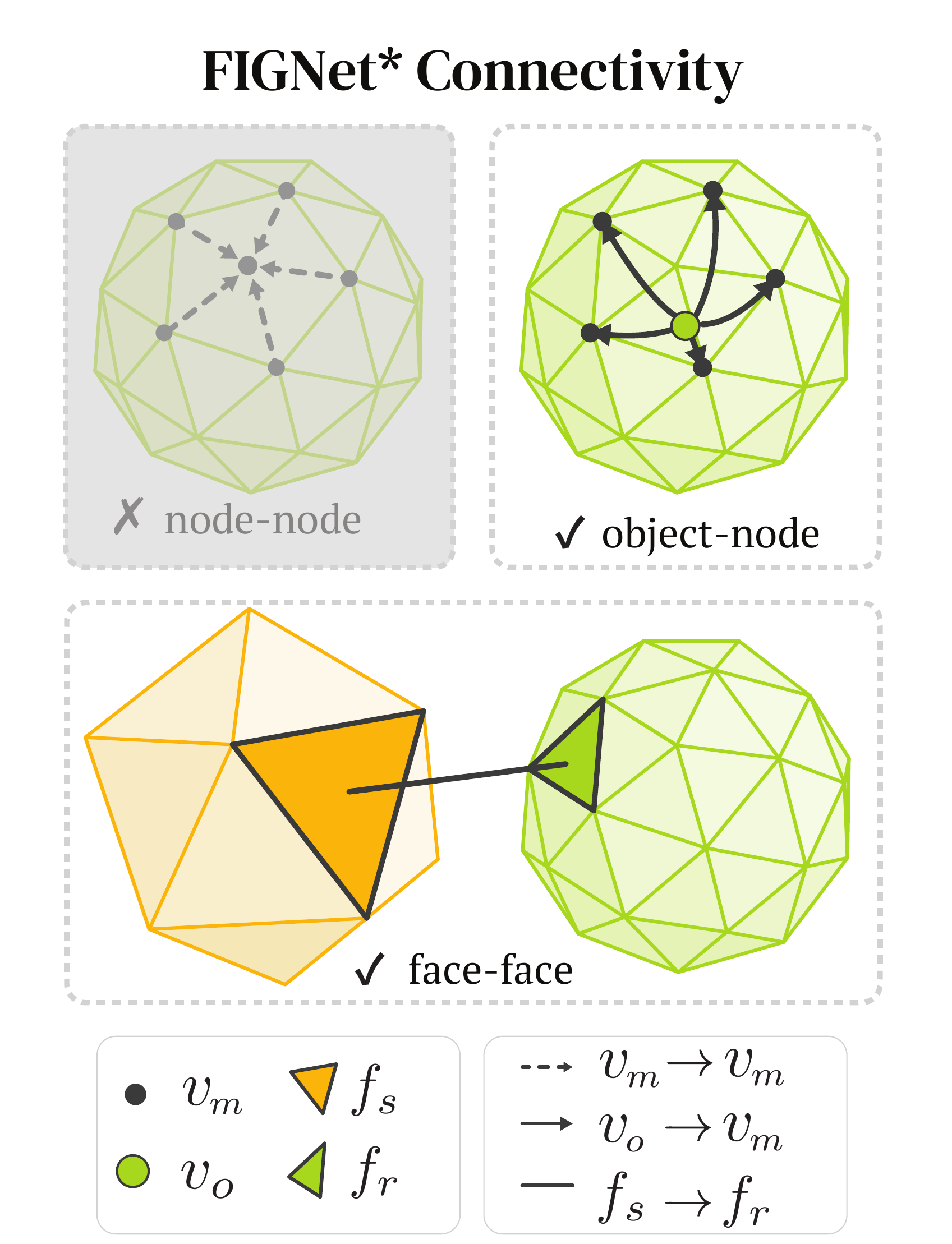}
\caption{\textbf{Architectural changes:} FIGNet* with respect to FIGNet.
}
\label{fig:method}
\vspace{-2.5em}
\end{wrapfigure}

FIGNet consists of two types of nodes (mesh nodes $\{\mathcal{V}_M \}$ and object nodes $\{\mathcal{V}_O \}$), and three types of bi-directional edges.

The mesh nodes $\{\mathcal{V}_M \}$ have input features  $\mathbf{v}^\text{M,features}_i = [\mathbf{x}^t_i-\mathbf{x}^{t-1}_i, \physprops_i, \kinematic_i, \mathbf{f}^t_i]$, where $\mathbf{x}^t_i$ is the position of the node at time $t$, $\physprops_i$ are static object properties like density and friction, $\kinematic_i$ is a binary ``static'' feature that indicates whether the node is subject to dynamics (e.g. the moving objects), or its position is set externally (e.g. the floor), and $\mathbf{f}^t_i = k_i ( \mathbf{x}^{t+1}_i-\mathbf{x}^{t}_i)$ is a feature that indicates how much kinematic nodes are going to move at the next time step. Object nodes $\{\mathcal{V}_O \}$ use the same feature description, with their positions $\mathbf{x}^t_i$ being the object's center of mass.

The three types of bi-directional edges include node-node, object-node, and face-face edges. Node-node edges $v_m \rightarrow v_m$ connect surface mesh nodes on a single object to one another. Object-node edges $v_o \rightarrow v_m$ connect object nodes $v_o$ to each mesh vertex $v_m$ of that object. Face-face edges connect faces on one sender object $f_{s}$ to another receiver object $f_{r}$. See Figure \ref{fig:method}.

Conceptually, the node-node edges enable the propagation of messages locally along an object's surface. However, in the case of rigid body collisions, collision information needs to be propagated instantaneously from one side of the object to the other, irrespective of the mesh complexity. Object-node edges enable this by having a single virtual object node $v_o$ at the center of each object which has bidirectional edges to each mesh node $v_m$ on the object's surface. Finally, to model the collision dynamics between rigid objects, face-face edges convey information about face interactions \textit{between} objects. FIGNet proposes a special hypergraph architecture for how to incorporate face-face edges into an Encode-Process-Decode graph network architecture. We defer further details of the FIGNet approach to \citep{allen2023fig}.

This approach works remarkably well for rigid body shapes but becomes intractably expensive as the complexity of each object mesh grows, since this will add a significant number of node-node (surface mesh) edges. Empirically, node-node edges often account for more than 50$\%$ of the total edges in FIGNet. FIGNet* makes a simple modification to FIGNet which removes the node-node (surface mesh) edges, keeping everything else identical. Surprisingly, this does not hurt the accuracy of FIGNet*, but dramatically improves memory and runtime performance for the rigid body settings examined in this paper. This works for rigid body dynamics because the \emph{collision edges} reason about the local geometry of two objects involved in contact, and this information can then be directly broadcasted to the whole shape using object-node edges.

This simple change to FIGNet unlocks the ability to train on much more complex scenes than was previously possible, as larger scenes fit into accelerator memory during training. We can therefore run FIGNet* on meshes extracted from real-world scenes, as well as simulations with more complex object geometries than previously possible.

\subsection{Connecting FIGNet* to Perception}
\label{sec:figreal}
\begin{figure}[h!]
\centering
\includegraphics[width=1.0\textwidth]{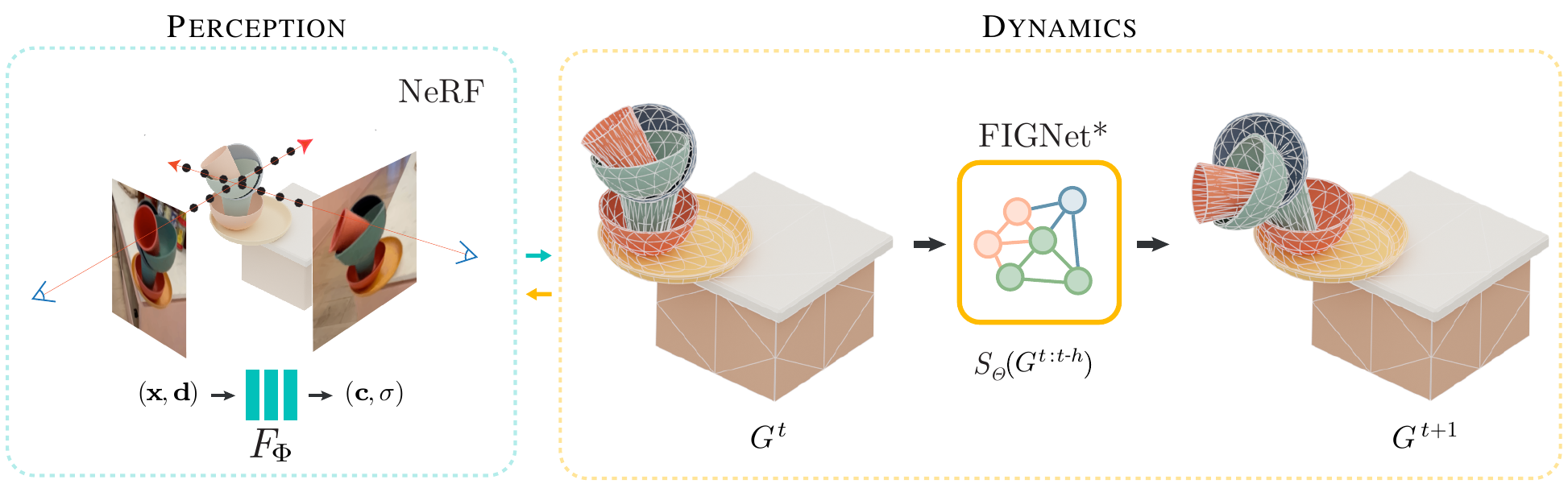}
\caption{\textbf{Perception Pipeline.} We demonstrate a two-way coupling approach, integrating FIGNet* with real-world scenes through NeRF. Initially, a static NeRF scene is trained using a collection of images capturing a real-world scene, enabling the extraction of the necessary meshes for FIGNet*. Upon obtaining the rollout trajectory, we derive a set of rigid body transformations, which are then utilized to edit the original NeRF. See \autoref{sec:figreal} for details.} 
\label{fig:overview}
\end{figure}

In this section we describe the procedure used to connect FIGNet* to the real world. We leverage Neural Radiance Fields (NeRFs) \citep{nerf, barron2022mipnerf360} as a perceptual front end to (1) extract the meshes required by FIGNet* for simulation and (2) re-render the scene with the transformations predicted by FIGNet* (\autoref{fig:overview}). This approach shares similarities with the method presented in \citep{qiao2023dmrf}, however,  here we demonstrate its implementation using a learned simulator.

\subsubsection{From NeRF to FIGNet*}

\paragraph{Learning a Neural Radiance Field:}
We first learn a NeRF from $W$ sparse input views $\{I\}_{1}^{W}$ and their associated camera intrinsics $\mathbf{K}$ and extrinsics. 
This representation models a view-dependent appearance function $F_\Phi$ that maps a 3D location $\posss=(x, y, z)$ and a viewing direction $\dirv$ to a radiance color $\colorv$ and a density $\density$.

\begin{equation}
    F_\Phi : (\posss, \dirv) \to (\colorv, \density)
\end{equation}

The geometries of all the objects in a scene represented by a NeRF are implicitly captured by $F_\Phi$. We only care about the density $\density$ for the geometry and can ignore the color $\colorv$ and the viewing direction $\dirv$. We slightly abuse the notation and define $\Fdensity(\posss) \to \density$ to denote the subpart of the NeRF that evaluates the density only.

\paragraph{Mesh Extraction:}
To extract the mesh of an individual object from the implicit function $\Fdensity$, we first need to define a volumetric boundary of the object.

We begin by generating $N$ binary segmentation masks, each capturing the object's shape from one of $N$ distinct viewpoints.
Each mask is created by calling XMEM \citep{cheng2022xmem} with the corresponding RGB image and a point prompt located at the center of the object. XMEM then identifies and labels all active pixels belonging to the object in each mask at the prompted location, resulting in a set of N segmentation masks $\{\mask\}_{1}^{N}$ that capture the object's shape from various perspectives. Empirically, we found that for simple objects like spheres, as few as two views from different angles are sufficient to accurately segment the object. However, one can use additional views for increased robustness or to capture finer details, particularly for more complex shapes.

We use the same procedure as described in \citep{cen2023segment} to unproject the pixels of the 2D masks into 3D points by leveraging the estimated depth $z(\mask)$ from the NeRF and the known camera intrinsics from which each mask was generated:

\begin{equation}
    \posss_\mask = z(\mask) * \mathbf{K}^{-1} \cdot (x(\mask), y(\mask), 1)^T
\end{equation}

The volumetric boundary $\boundary \in \mathbb{R}^{2 \times 3}$ can be then obtained as

\begin{equation}
    \boundary = \{\min(\posss_\mask), \max(\posss_\mask)\}_1^{N}
\end{equation}

To extract the mesh of the object $M_o$ within the volume $\boundary$, we employ the Marching Cubes algorithm (\texttt{m\_cubes}) \citep{marching_cubes}. This algorithm uses samples of the density field from a regular grid of J points inside the boundary $\posss_j \in \boundary$ as $\mathbf{\density_o} = \{\Fdensity(\posss_j)\}_{1}^{J}$ and a threshold value $\sigma_{thrs}$.
To manage the potentially high number of vertices and faces in the generated mesh, we perform an additional decimation step (\texttt{decimate}). We employ the Quadric Error Metric Decimation method by Garland and Heckbert \citep{mc}. This technique preserves the primary features of the mesh while allowing us to control the final mesh complexity through a user-specified target number of faces $n_{f}$.

\begin{equation}
    M_o = \texttt{decimate}(\,
        \texttt{m\_cubes}(\mathbf{\density_o}, \density_{thrs}), \,n_{f})
\end{equation}

\paragraph{Building the Graph} To specify the object whose motion we want to simulate, we define the mesh $M_o$ as the active object in the graph, with all other objects considered static. 
We then repeat the same mesh extraction procedure described above on an offset version of the scene volume $(\boundary - \boundaryShift)$ to obtain the passive mesh $M_{passive}$ representing the static environment with $\kinematic_i$ set to True. Both meshes are used to construct the initial graph $\graphT{t}$ for FIGNet and FIGNet*. 
We do not infer static properties like mass, friction, elasticity, etc for meshes extracted from the scene. Instead we use the default parameters provided in \autoref{tab:parameters}. Future work will be needed to infer these properties from object dynamics.

We generate the history $\graphT{t-1}$ using the same mesh but shifted downwards by a $\Delta z$ amount twice to simulate an object being dropped vertically. 

\subsubsection{From FIGNet* to NeRF}

We obtain a rollout trajectory by iteratively applying FIGNet* over $T$ time steps. Starting from the initial graph and its history to obtain $(\graphT{t+1}, \graphT{t+2}, \cdots, \graphT{t+T})$. This can be equivalently seen as a sequence of rigid transformations $(\transformK{t+1}, \transformK{t+2}, \cdots, \transformK{t+T})$ that are applied to $M_o$.

Given the bounding volume of each object $\boundary$ and a rigid transformation $R_t$ at time $t$, we can reuse the static NeRF function $\F$ to render the rollout by editing the original static NeRF described by $\F$ via ray bending \citep{nerfshop}. We restrict the bending of the ray $b$ to be the rigid transformation returned by FIGNet* as

\begin{equation}
    \Fedit : (\, b(\posss, \transformK{t} \,), \dirv) \to (\colorv, \density),
\end{equation}

where $b(\posss, \transformK{t} \,)$ can be either

\begin{equation}
    b_{move}(\posss, \transformK{t}) = 
     \begin{cases}
       \transformK{t} \times \posss &\quad\text{if } \posss \in \boundary\text{,}\\
    (\transformK{t})^{-1} \times \posss &\quad\text{if } \posss \in \transformK{t} \times \boundary \text{,}\\
       \posss &\quad\text{otherwise.} \\ 
     \end{cases}
\end{equation}

or

\begin{equation}
    b_{duplicate}(\posss, \transformK{t}) = 
     \begin{cases}
    (\transformK{t})^{-1} \times \posss &\quad\text{if } \posss \in \transformK{t} \times \boundary \text{,}\\
       \posss &\quad\text{otherwise.} \\ 
     \end{cases}
\end{equation}

meaning that the active object has the option to be either moved or copy-pasted during the rollout.

We then generate the final sequence of rollout images from a chosen viewpoint 
$\hat{\dirv}$ across all time steps. This involves applying NeRF's classic volume rendering pipeline with the transformed radiance field $\Fedit$ incorporating object movement. At each step, we adjust the radiance field based on the applied rigid transformation, effectively capturing the dynamic appearance of the object throughout the rollout sequence $\{\Fedit( b(\posss, R_t \,), \hat{\dirv})\}_{t=1}^{k}$.
\section{Results}
We test FIGNet* on both simulated and real data. In simulation, we show that FIGNet* outperforms FIGNet in memory consumption and runtime while maintaining accuracy for a standard rigid body dynamics benchmark \citep{greff2022kubric}. For real data, we show that FIGNet* can be run on views of real scenes collected from multiple cameras, making plausible trajectories despite training in simulation on perfect state information.

\subsection{Simulation}
For our simulation results, we use the MOVi-B and MOVi-C Kubric datasets \citep{greff2022kubric}. In both setups, multiple rigid objects are tossed together onto the floor using the PyBullet \citep{Coumans} simulator to predict trajectories. MOVi-B consists of scenes involving 3-10 objects selected from 11 different shapes being tossed. The shapes include teapots, gears, and torus knots, with a few hundred up to just over one thousand vertices per object. MOVi-C consists of scenes involving 3-10 objects selected from 1030 different shapes taken from the Google Scanned Objects dataset \citep{downs2022google}. MOVi-C shapes tend to be more complex than MOVi-B shapes, and have up to several thousand or tens of thousands of vertices.

We report four metrics in Table \ref{tab:kubric_compare}: peak memory consumption, runtime per simulation step, translation error, and rotation error. Translation and rotation root-mean-squared error (RMSE) are calculated with respect to the ground truth state after 50 rollout steps.
\begin{table}[h]
 \centering
 \caption{Comparison metrics for FIGNet and FIGNet* on Kubric MOVi-B and MOVi-C}
 \small
\begin{tabular}{c|c|c|c|c|c|c}

Dataset & Model & Memory (MiB) & Runtime (ms) & Trans. Err. (m) & Rot. Err. (deg) & Edge Count ($\#$)\\
\toprule
\multirow{2}{*}{MOVi-B} & FIGNet & 63.38 $\pm$ 3.32 & 26.38 $\pm$ 0.73 & 0.14 $\pm$ 0.01 & 14.99 $\pm$ 0.67 & 24514 $\pm$ 906 \\
& FIGNet* & \textbf{50.08 $\pm$ 3.37} & \textbf{19.41 $\pm$ 0.24} & 0.13 $\pm$ 0.01 & 15.96 $\pm$ 0.87 & \textbf{8630 $\pm$ 714} \\
\midrule
\multirow{2}{*}{MOVi-C} & FIGNet & OOM & -- & -- & -- & -- \\
& FIGNet* & \textbf{71.79 $\pm$ 6.39} & \textbf{20.42 $\pm$ 0.64} & \textbf{0.18 $\pm$ 0.01} & \textbf{19.82 $\pm$ 0.64} & \textbf{11401 $\pm$ 975} \\
\end{tabular}
\label{tab:kubric_compare}

\end{table}

\begin{figure}
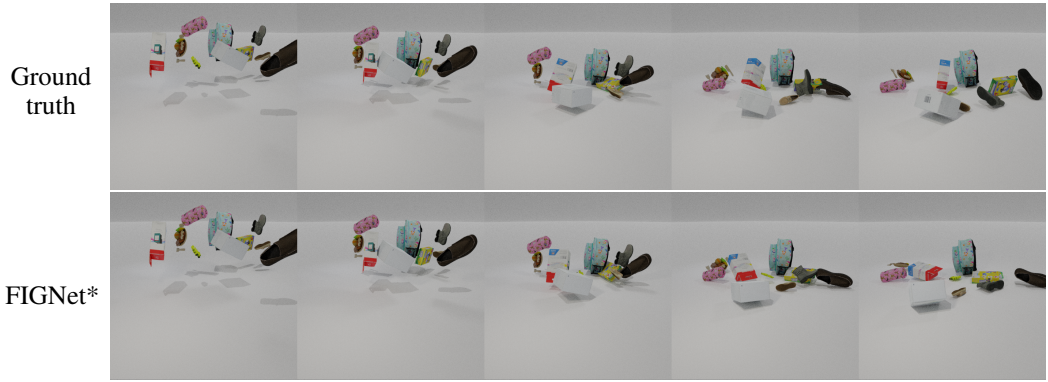

\centering
\MOVICROW{023}
\caption{\textbf{Qualitative results for simulation.} FIGNet* rollout for complex MOVi-C simulation  which could not be represented in memory for FIGNet.}
\label{fig:movic_rollout}
\end{figure}

For MOVi-B, FIGNet* matches FIGNet's performance in translation and rotation error, performing slightly better in translation, and slightly worse on rotation. However, FIGNet* uses significantly less memory than FIGNet while also having a 20$\%$ faster runtime. These differences in memory consumption and runtime allow us to train FIGNet* on the much more complex MOVi-C dataset (example trajectory in Figure \ref{fig:movic_rollout}), which causes OOM errors when attempting to train FIGNet even with 16 A100 GPUs. On MOVi-C, the memory consumption is higher, but runtime remains almost as fast. Similarly, since MOVi-C is more complex than MOVi-B, the translation and rotation errors for FIGNet* are higher, but not significantly so.

Overall, this suggests that FIGNet* is a viable alternative to FIGNet. It maintains accuracy while significantly reducing memory consumption and runtime, allowing us to train FIGNet* on more complex datasets than can be fit into FIGNet memory.


\subsection{Real world}
We present our results on linking FIGNet* with real-world scene inputs. Note that this is a proof-of-concept only, that is we do not compare to real ground truth dynamics, instead leaving that for future work. For comparisons between FIGNet and FIGNet* on real data, FIGNet models were trained in simulation on Kubric MOVi-B, while FIGNet* models were trained in simulation on Kubric MOVi-C.

For our real-world results, we used two scenes: our custom-made \textsc{kitchen} scene filled with common elements such as fruits and baskets (See \autoref{ap:real} for details), the \textsc{garden}-outdoor and \textsc{kitchen counter}-indoor scenes introduced in \citep{barron2022mipnerf360} and the \textsc{figurines} scene introduced in \citep{lerf2023}. These scenes consist of 360-degree image sets captured with different cameras. We used a MipNerf360 \citep{barron2022mipnerf360} implementation for the NeRF front end.


\begin{figure}[ht]
\centering
\includegraphics[width=1.0\textwidth]{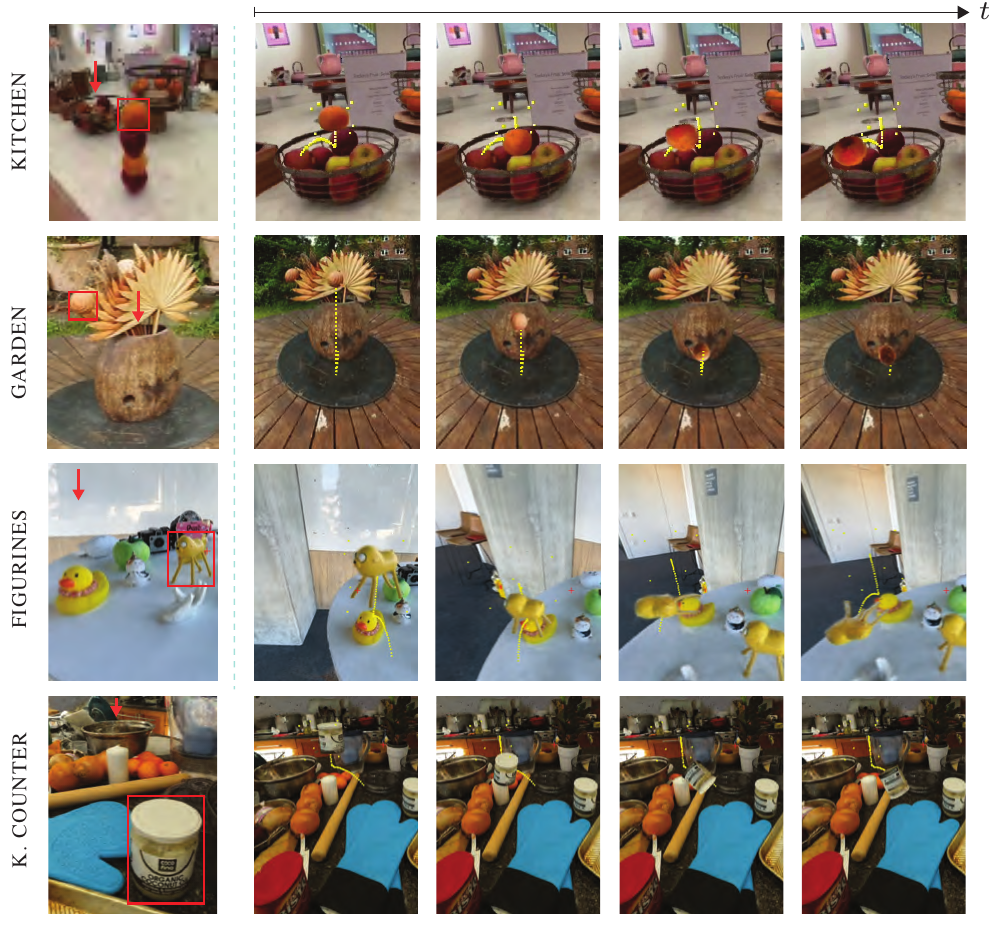}
\caption{\textbf{Qualitative results for real world scenes.} \textit{Left:} Initial NeRF rendering of the static real-world scene. The desired active object is outlined in red, with a red arrow indicating its intended starting position.  \textit{Right:}  FIGNet* rollouts simulating the object's motion for $k=30$ time steps (rendered from a different viewpoint) after being dropped from the initial position. The complete trajectory is traced in yellow. Here we used $b_{duplicate}$ as the ray bending function meaning the active object is copy pasted into the starting position at the beginning of the rollout (See the website for videos and \autoref{ap:real:segmentation} for details on the mesh extraction procedure described in \autoref{sec:figreal}). 
}
\label{fig:real_results}
\end{figure}

\paragraph{Qualitative Results.} We show qualitative
FIGNet* rollouts on both real world scenes using the full pipeline described in \autoref{sec:figreal}. For all the scenes, we manually selected 2 views of the active object (highlighted in the red boxes) to compute the bounding volume $\boundary$ and the subsequent mesh $M_o$ (See \autoref{ap:real:segmentation}). By creating a history based on downward vertical displacement of the chosen mesh, we are effectively simulating a motion similar to dropping. Figure \ref{fig:real_results} illustrates the bouncing behaviors of various objects falling onto other objects. Note the sharp rotation of the orange at the end of the bounce (last frame) in the \textsc{kitchen} scene, and how rendering with the transformed $\Fedit$ works when the orange is flipped upside down. We can observe similar results for the \textsc{figurines} scene, where we selected two views of the dog figurine with long thing legs and simulate a dropping motion onto a duck. Our perception pipeline can realistically simulate and re-render the dropping motion of objects captured within these real scenes by reusing the static NeRF scene with the FIGNet* transformations \footnote{See \url{https://sites.google.com/view/fignetstar/} for videos.}.

\begin{figure}[ht!]
\centering
\includegraphics[width=0.85\textwidth]{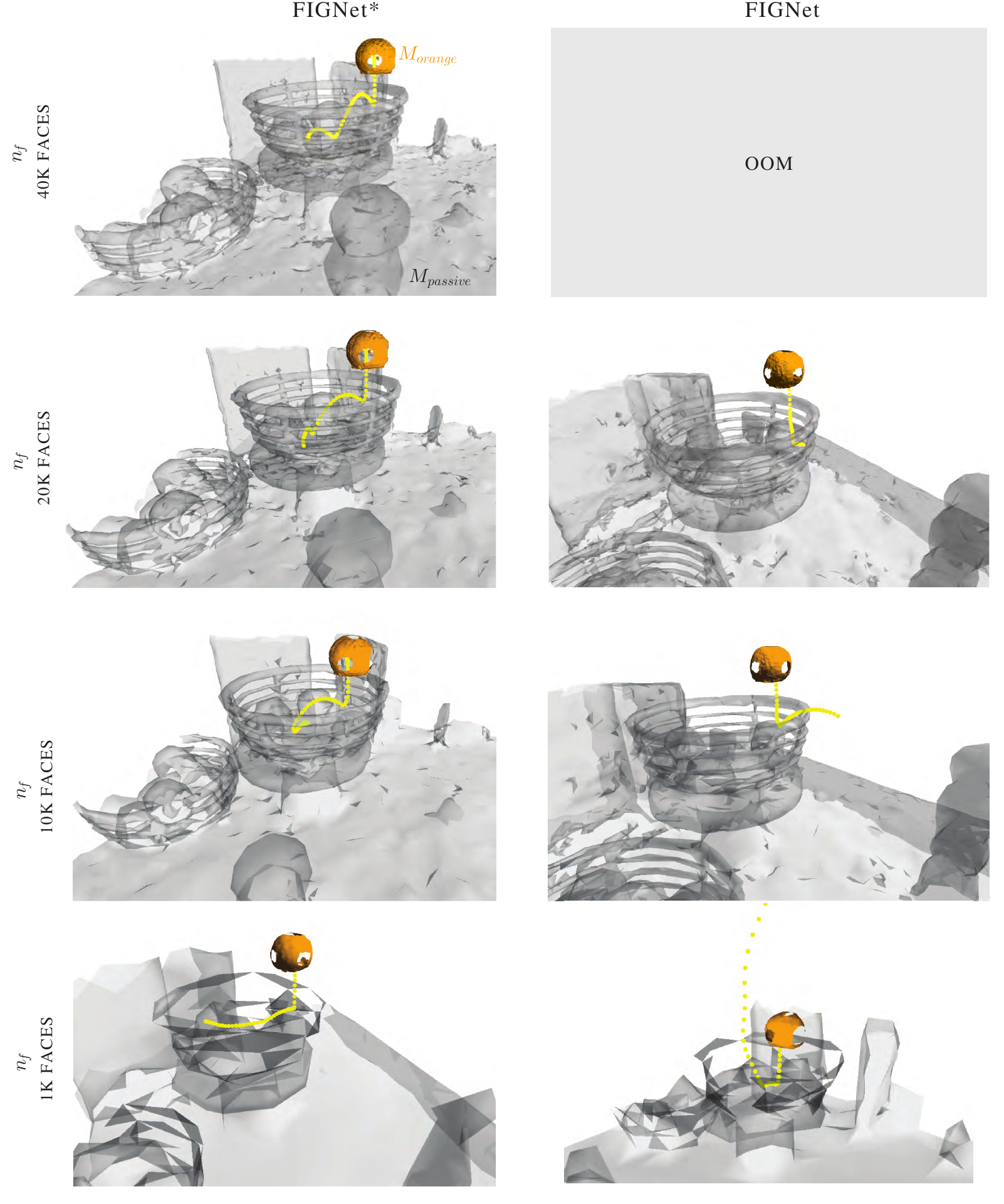}
\caption{
FIGNet and FIGNet* comparison for different levels of decimation: High-quality meshes lead to out-of-memory issues on FIGNet, while lower resolutions result in implausible trajectories (e.g., orange penetrating the basket). Notably, FIGNet*'s performance gracefully degrades with mesh quality, indicating enhanced robustness and memory efficiency. The gray mesh depicts the passive object, and the colored mesh corresponds to the active object.
}
\label{fig:real:decimation}
\end{figure}

\paragraph{Effect of decimation.} 

The marching cube algorithm often results in oversampled meshes characterized by an elevated node count. While the implementation of a controllable parameter for mesh decimation ($n_f$) is an effective strategy to address this challenge, it is important to note that the extent of decimation can adversely affect the quality of simulations, especially in cases involving complex geometries. The advantage of using FIGNet* lies in its reduced memory requirements, which permits a less rigorous decimation process in comparison to FIGNet. To demonstrate this, we simulated a scene with two distinct levels of decimation  (\autoref{fig:real:decimation}). This experiment highlights instances where FIGNet's memory capacity is exceeded, showcasing the benefits of FIGNet* in such scenarios.


\paragraph{Effect of perception noise.} 
Real-world meshes extracted from pipelines like NeRF, primarily optimized for rendering quality, often exhibit noise and imperfections (\autoref{fig:real:noisy}). Unlike the clean training data used for FIGNet* and FIGNet, these meshes are far from ideal. Nevertheless, both models can successfully handle rollouts even with such challenging real-world data.

\section{Discussion}
We showed that a surprisingly simple modification to FIGNet, the removal of the surface mesh edges, allowed us to create a model with low enough memory consumption to support training on unprecedentedly complex scenes. This unlocked the ability to interface FIGNet* with real world scenes by using a combination of Neural Radiance Fields (NeRFs) and object selection (XMem) to convert real scenes into object-based mesh representations. In combination with volumetric NeRF editing, this allowed us to simulate videos of alternative physical futures for real scenes.

We believe that this explicitly 3D approach to video editing and generation has significant promise for robotics and graphics applications. It allows a model to be pre-trained from simulation data, while still generalizing to real scenes. FIGNet* generalizes surprisingly well to noisy meshes extracted from NeRFs, especially considering that it was trained in simulation with nearly perfect state information (positions, rotations, and shapes of objects). We imagine that this approach could further support future applications including ``virtualization'' of real scenes, where users may be interested in editing those scenes and simulating possible future outcomes.

There are many exciting directions for future work with FIGNet*. In particular, while fine-tuning a pre-trained FIGNet* model to a real video was outside the scope of this paper, we believe this is a natural next step.
Since FIGNet* is entirely composed of neural networks, fine-tuning from real world dynamics directly into the weights of FIGNet* could be a viable alternative to system identification for robotics. Future work will be needed to determine the details of how to perform fine-tuning in a data efficient manner.

\bibliography{main}
\bibliographystyle{iclr2024_conference}

\newpage
\appendix
{\Large Appendix}
\vspace{0.5cm}

\section{Decimation experiments}
\label{ap:real:decimation}

We qualitatively evaluated the impact of mesh decimation on rollouts for FIGNet and FIGNet* in the \textsc{kitchen} scene (\autoref{fig:real:decimation}). With higher quality meshes (lower decimation), FIGNet tends to run out of memory, whereas lower quality meshes (higher decimation) often result in unrealistic rollouts. In such cases, objects (orange) may pass through solid objects (basket), as observed with meshes of 1k faces. In contrast, FIGNet*'s rollout trajectories exhibit a graceful degradation with increased levels of decimation, maintaining relative stability even at very high decimation levels ($n_f=1000$, which means approximately 1\% of the original faces are preserved)

\begin{figure}[ht]
\centering
\includegraphics[width=0.9\textwidth]{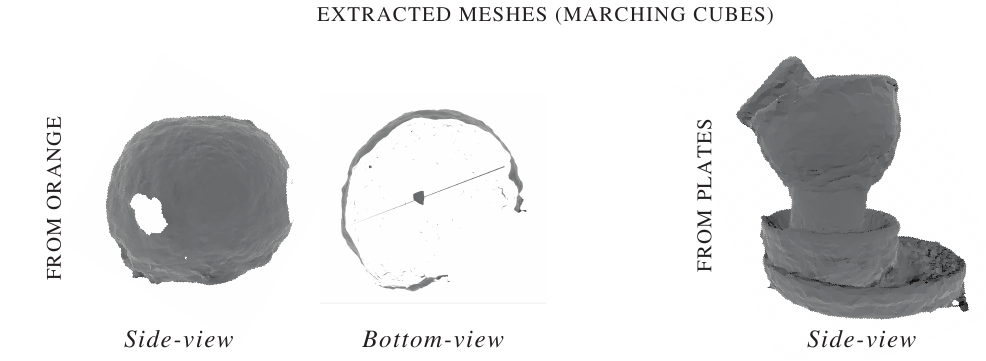}
\caption{Noisy meshes extracted from NeRF, including the orange object on the left missing its bottom face (from \autoref{fig:real_results}) and the plates (from \autoref{fig:real:more}). Notably, both FIGNet and FIGNet* can handle rollouts even with such mesh imperfections, demonstrating their robustness to real-world data challenges.}
\label{fig:real:noisy}
\end{figure}

\begin{figure}[ht]
\centering
\includegraphics[width=0.85\textwidth]{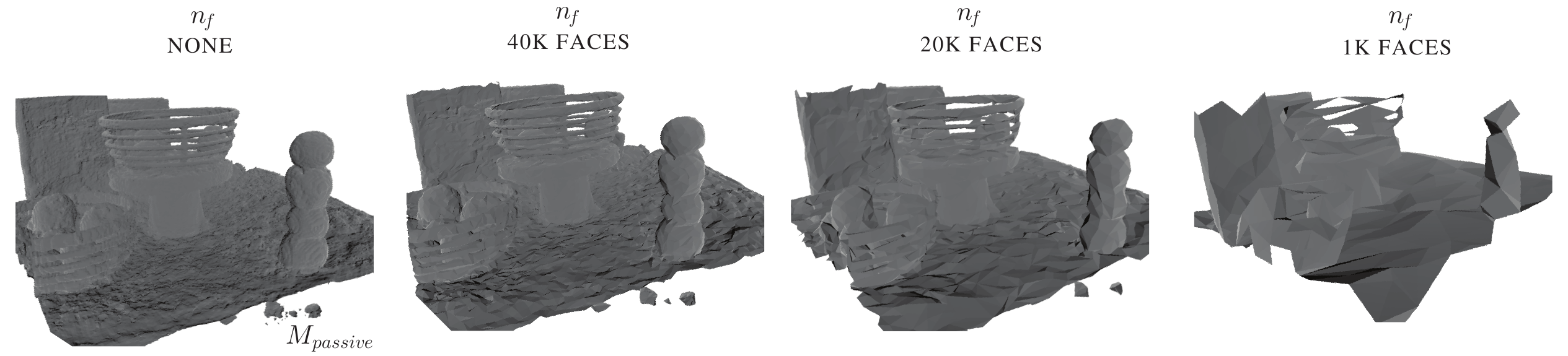}
\caption{
Effect of the decimation parameter on the mesh quality. \textit{Left:} no decimation. \textit{Right: } high decimation.
}
\label{fig:real:decimation:meshes}
\end{figure}






\section{Image segmentations}
\label{ap:real:segmentation}

\begin{figure}[ht]
\centering
\includegraphics[width=0.99\textwidth]{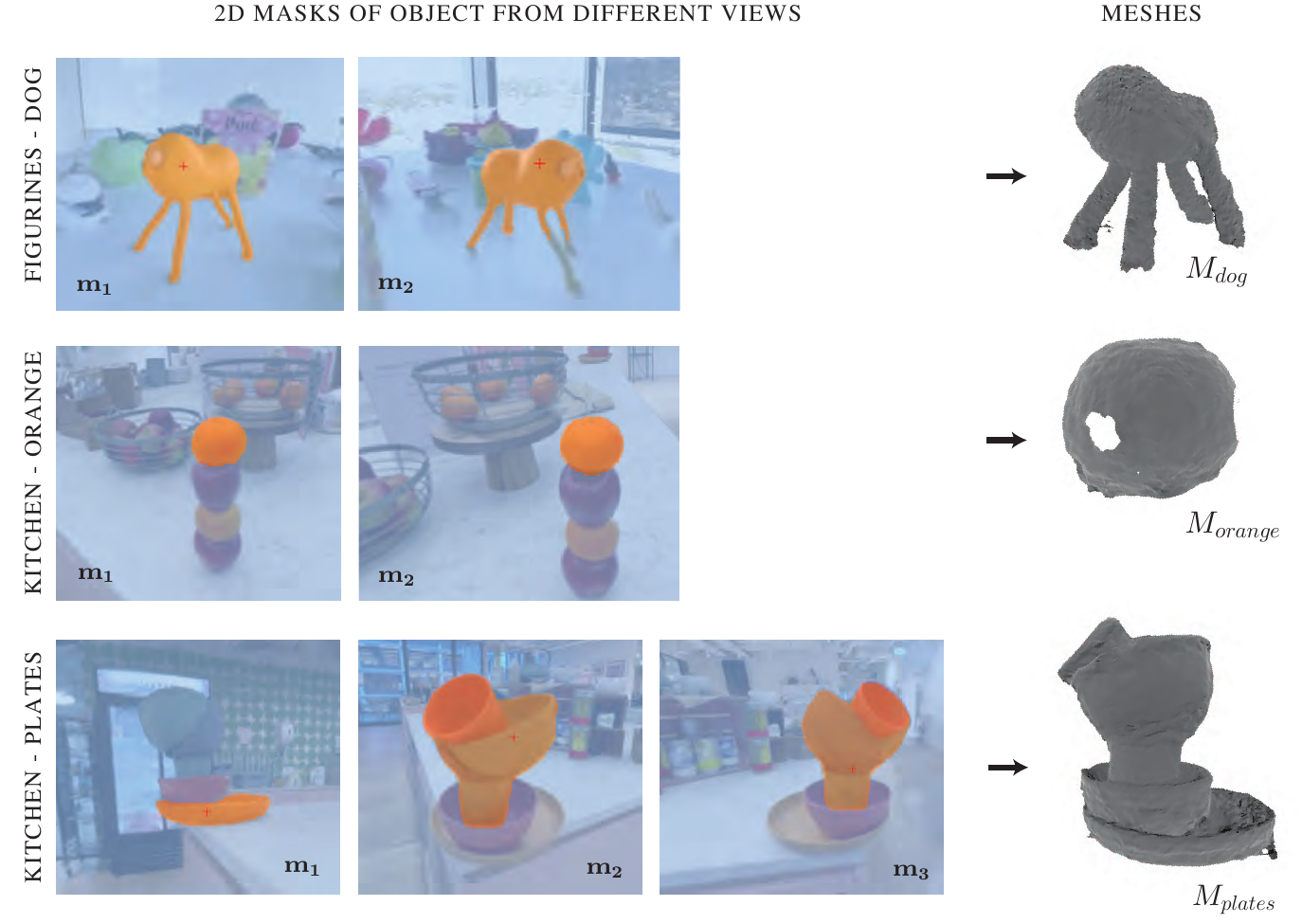}
\caption{\textit{Left:} Selected views to generate the objects masks for the \textsc{figurines} and \textsc{kitchen} scenes. The top row corresponds to the rendered image in RGB with each orange mask $\{\mask\}_{1}^{N}$ (overlaid in light orange) obtained by XMEM's \citep{cheng2022xmem}. The bottom row illustrates the same procedure for the plates on the same scene. Note that partial segmentations from different views can also be used to build the volumetric boundary of the object. \textit{Right:} the obtained mesh $M_o$ from each of the masks after decimation.}
\label{fig:real:segmentations}
\end{figure}

\begin{figure}[ht]
\centering
\includegraphics[width=0.85\textwidth]{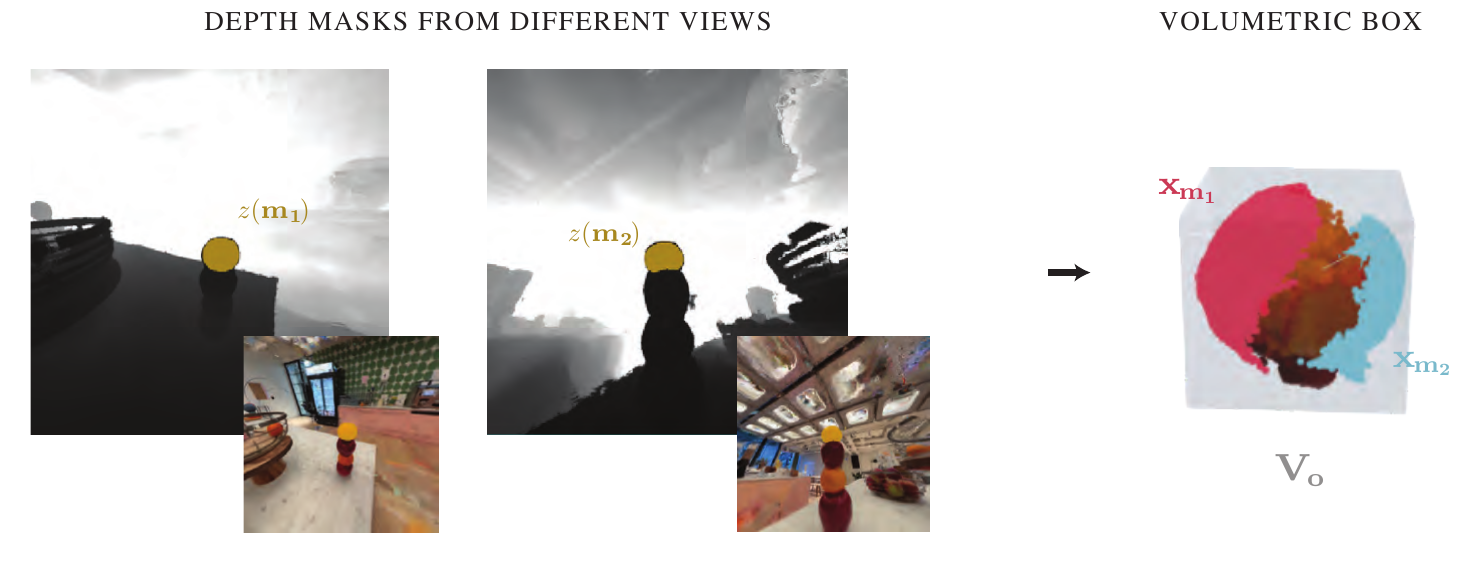}
\caption{Visualizing the generation of the orange's volumetric box from depth masks in the \textsc{kitchen} scene.}
\label{fig:real:bbox}
\end{figure}

We provide some examples of how the mesh extraction procedure described in \autoref{sec:figreal} works in \autoref{fig:real:segmentations} and \autoref{fig:real:bbox}.  

\section{Kitchen scene details}
\label{ap:real}
\begin{figure}[h]
\centering
\includegraphics[width=0.85\textwidth]{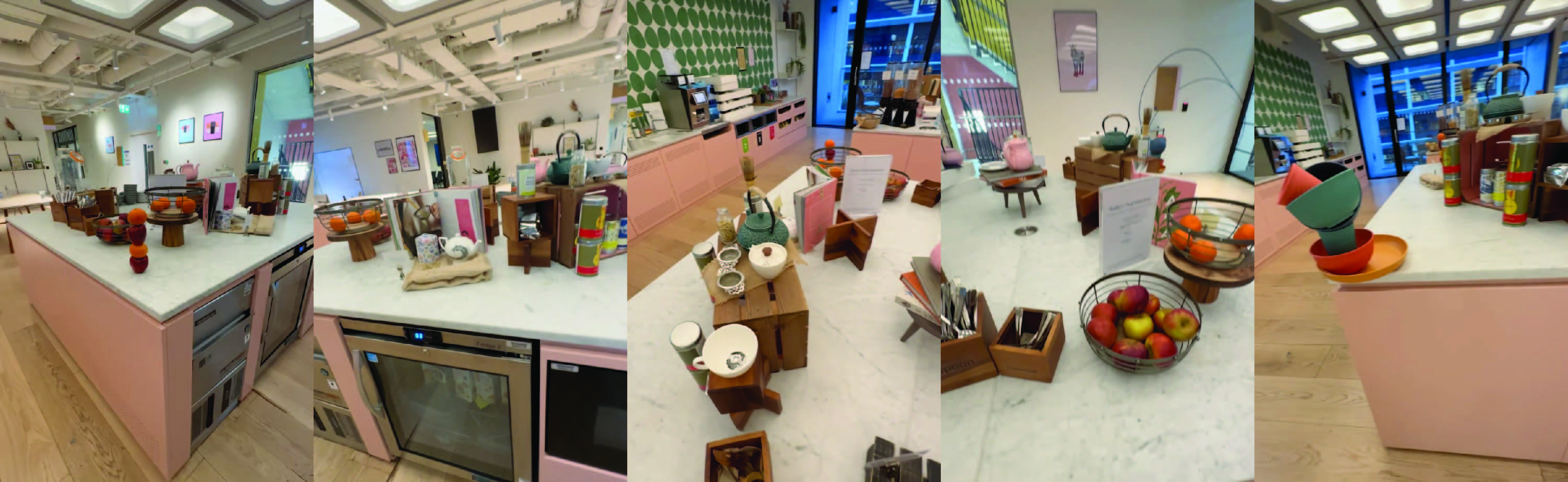}
\caption{
Example frames from the \textsc{kitchen} scene video.
}
\label{fig:real:images}
\end{figure}

We collected 1027 images of a \textsc{kitchen} scene that included different elements such as apples, oranges, baskets and plates. We extracted the images from a video recorded with an iPhone 14 Pro at 60fps and HEVC format (\autoref{fig:real:images}). We used COLMAP \citep{colmap} to estimate the camera poses from the images.

\section{Implementation details} \label{sec:implementation_details}

\subsection{Hyper-parameters}
\label{ap:params}
FIGNet* is trained identically to FIGNet \citep{allen2023fig}. 

\paragraph{MLPs for Encoder, Processor, Decoder} We use MLPs with 2 hidden layers, and 128 hidden and output sizes (except the decoder MLP, with an output size of 3). All MLPs, except for those in the decoder, are followed by a LayerNorm\citep{ba2016layer} layer.

\paragraph{Optimization} All models are trained to 1M steps with a batch size of 128 across 8 TPU devices. We use Adam optimizer, and an an exponential learning rate decay from 1e-3 to 1e-4.

\begin{table}[h] 
 \centering
 \caption{NeRF Training Parameters}
\begin{tabular}{c|c|c}

Type & Parameter & Value \\
\toprule
General & near & 0. \\
General & far & 1e6 \\
General & lr\_delay\_steps & 100 \\
General & batch\_size & 65536 \\
General & lr\_init & 1e-2 \\
General & lr\_final & 1e-3 \\
General & adam\_beta1 & 0.9 \\
General & adam\_beta2 & 0.99\\
General & adam\_eps & 1e-15\\
General & cast\_rays\_in\_eval\_step & True \\
General & cast\_rays\_in\_train\_step & True \\
General & num\_glo\_features & 4 \\
Model & sampling\_strategy & ((0, 0, 64), (0, 0, 64), (1, 1, 32)) \\
Model & grid\_params\_per\_level & (1, 4) \\
Hash & hash\_map\_size & 2097152 \\
Hash & scale\_supersample & 1. \\
Hash & max\_grid\_size & 8192 \\
MLP & net\_depth & 1 \\
MLP & net\_width & 64 \\
MLP & disable\_density\_normals & True \\
MLP & density\_activation & @math.safe\_exp \\
MLP & bottleneck\_width & 15 \\
MLP & net\_depth\_viewdirs & 2 \\
MLP & net\_width\_viewdirs & 64 \\
\end{tabular}
\label{tab:kubric_compare}
\end{table}

\begin{table}[h] 
 \centering
 \caption{Default Physical Parameters}
\begin{tabular}{c|c|c|c|c}
Model & Type & Mass & Friction & Restitution \\
\toprule
FIGNet* & Active &  1e-3 & 0.5 & 0.5 \\
FIGNet* & Passive & 0 & 0.5 & 0.3 \\
\midrule
FIGNet & Active &  1.0 & 0.8 & 0.7 \\
FIGNet & Passive & 0 & 0.5 & 0.3 \\
\end{tabular}
\label{tab:parameters}
\end{table}

\section{Additional rollouts for the real world scenes}

\begin{figure}[ht]
\centering
\includegraphics[width=1.0\textwidth]{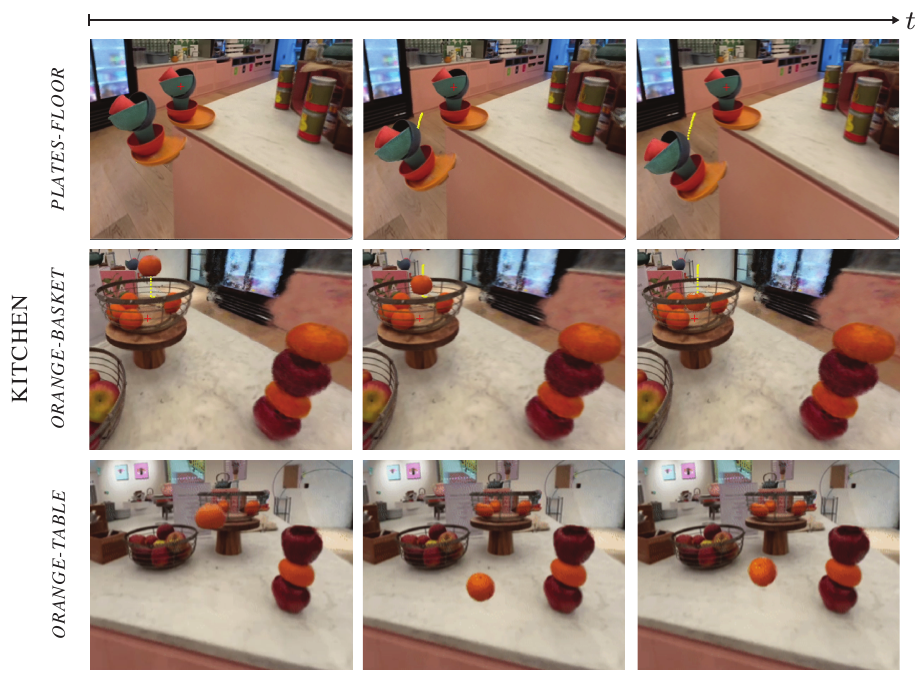}
\caption{Additional examples of FIGNet* rollouts on the \textsc{kitchen} scene. The final row was generated using the $b_{move}$ ray bending function (moving the orange from the fruit tower to the starting position), while the other rows used 
$b_{duplicate}$ (copy-pasting the object).}
\label{fig:real:more}
\end{figure}

We provide additional rollout examples for the \textsc{kitchen} scene in \autoref{fig:real:more} and in the website. 
\textsc{plates-floor:} Duplicating the stack of plates on the right and shifting their initial position to the left. \textsc{orange-basket:} The top orange from the stack of fruits is duplicated and dropped on top of a basket of oranges. Note the correct depth ordering of the orange with respect to the basket. \textsc{orange-table}: the orange is dropped on the table. 

\section{Example rollouts for MOVi-C}

Additional simulation rollouts of FIGNet* on Kubric MOVI-C.

\begin{figure*}[ht!]
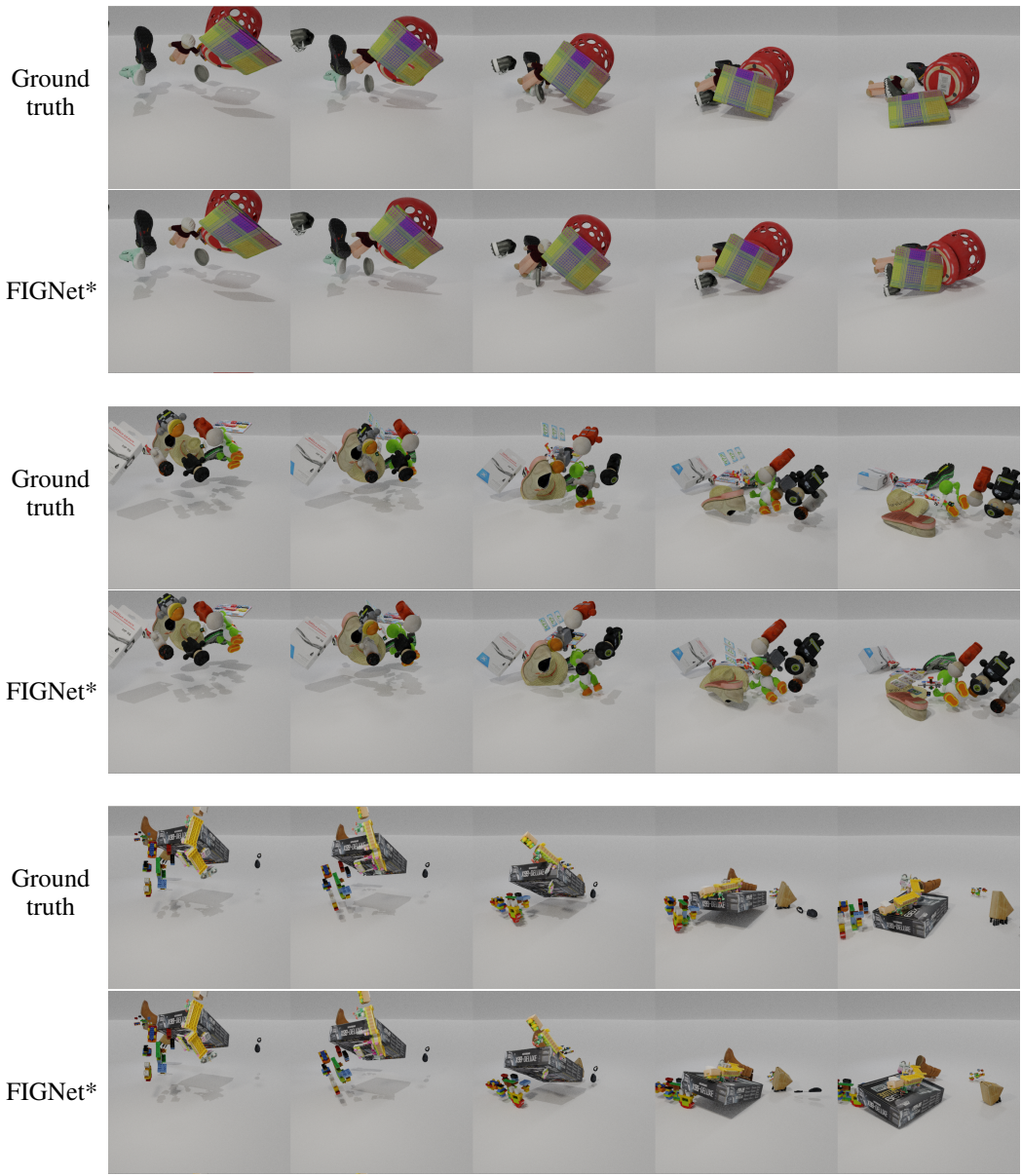

\centering
\MOVICROW{\movicrolloutone}
\bigskip
\MOVICROW{\movicrollouttwo}
\bigskip
\MOVICROW{\movicrolloutthree}
\caption{Rollout of FIGNet* Kubric MOVi-C.}
\label{fig:rollouts}
\end{figure*}

\end{document}